\documentclass[10pt,twocolumn,letterpaper]{article}

\usepackage[final]{cvpr}      

\usepackage{graphicx}
\usepackage{amsmath}
\usepackage{amssymb}
\usepackage{booktabs}
\usepackage{array}
\newcommand \footnoteONLYtext[1]
{
	\let \mybackup \thefootnote
	\let \thefootnote \relax
	\footnotetext{#1}
	\let \thefootnote \mybackup
	\let \mybackup \imareallyundefinedcommand
}
\usepackage{times}  
\usepackage{helvet} 
\usepackage{courier}  
\usepackage{algorithm}
\usepackage{algorithmic}
\usepackage{amssymb}
\usepackage{amsfonts}
\usepackage{amsthm,amsmath,amssymb}
\usepackage{booktabs}
\usepackage{bm}
\usepackage{caption,subcaption}
\usepackage{color}
\usepackage{epsfig}
\usepackage{multirow}
\usepackage{multicol}
\usepackage{mathrsfs}
\usepackage{pifont}
\usepackage{textcomp}
\usepackage{threeparttable}
\usepackage{url}
\usepackage{xcolor}
\usepackage{ulem}

\usepackage[pagebackref,breaklinks,colorlinks]{hyperref}
\usepackage[capitalize]{cleveref}

\crefname{section}{Sec.}{Secs.}
\Crefname{section}{Section}{Sections}
\Crefname{table}{Table}{Tables}
\crefname{table}{Tab.}{Tabs.}

\def\cvprPaperID{} 
\def\confName{CVPR}
\def\confYear{2022}

\begin{document}

\title{Domain Invariant Masked Autoencoders for Self-supervised Learning from Multi-domains}

\author{Haiyang Yang$^{1,4\ast}$,Meilin Chen$^{2,4\ast}$,Yizhou Wang$^{2,4\ast}$, Shixiang Tang$^{3\ast}$, Feng Zhu$^{4}$, \\
Lei Bai$^{3}$, Rui Zhao$^{4,5}$, Wanli Ouyang$^3$\\
$^1$WuHan University, $^2$Zhejiang University, $^3$The University of Sydney,$^4$SenseTime Research,\\ $^5$Qing Yuan Research Institute, Shanghai Jiao Tong University, Shanghai, China\\
{\tt\small yanghaiyang@sensetime.com, \{yizhouwang, merlinis\}@zju.edu.cn, stan3906@uni.sydney.edu.au,}\\
{\tt\small \{zhufeng, zhaorui\}@sensetime.com, baisanshi@gmail.com, wanli.ouyang@sydney.edu.au}}
\maketitle

\begin{abstract}
  Generalizing learned representations across significantly different visual domains is a fundamental yet crucial ability of the human visual system. While recent self-supervised learning methods have achieved good performances with evaluation set on the same domain as the training set, they will have an undesirable performance decrease when tested on a different domain. Therefore, the self-supervised learning from multiple domains task is proposed to learn domain-invariant features that are not only suitable for evaluation on the same domain as the training set, but also can be generalized to unseen domains. In this paper, we propose a Domain-invariant Masked AutoEncoder (DiMAE) for self-supervised learning from multi-domains, which designs a new pretext task, \emph{i.e.,} the cross-domain reconstruction task, to learn domain-invariant features. The core idea is to augment the input image with style noise from different domains and then reconstruct the image from the embedding of the augmented image, regularizing the encoder to learn domain-invariant features. To accomplish the idea, DiMAE contains two critical designs,  
1) content-preserved style mix, which adds style information from other domains to input while persevering the content in a parameter-free manner, and 2)
multiple domain-specific decoders, which recovers the corresponding domain style of input to the encoded domain-invariant features for reconstruction.
Experiments on PACS and DomainNet illustrate that DiMAE achieves considerable gains compared with recent state-of-the-art methods. Code will be released upon acceptance.

\end{abstract}

\footnoteONLYtext{$\ast$ The work was done during an internship at SenseTime.}

\section{Introduction}
Recent advances on self-supervised learning~(SSL) with the contrastive loss~\cite{he2020momentum,chen2020simple,chen2020improved,tian2020makes} have shown to be effective in easing the burden of manual annotation, and achieved comparable performance with supervised learning methods. When trained on large-scale datasets, \emph{e.g.} ImageNet~\cite{deng2009imagenet}, self-supervised learning methods are capable of learning high-level semantic image representations~\cite{zbontar2021barlow,ericsson2020how,zhao2021what} that are transferable to various downstream tasks without using expensive annotated labels. However, the great success of existing self-supervised learning methods implicitly relies on the assumption that training and testing sets are identically distributed, and thus these methods will suffer an undesirable performance drop when the trained model is tested on other domains~\cite{zhuang2020comprehensive,sariyildiz2021concept,wang2021revisiting} that do not exist in the training set.

Self-supervised learning from multi-domain data aims at learning domain invariant representations that are not only suitable for domains in the training set, but also can generalize well to other domains missing in the training set. Existing methods can be generally divided into two categories, \emph{i.e,} self-prediction methods and contrastive-based methods. Early methods for self-supervised learning from multi-domain data append self-prediction tasks to learn domain-invariant features. For example,~\cite{feng2019self} randomly rotates the input image and regularizes the model to predict the rotation angle~\cite{gidaris2018unsupervised} to increase the model generalization ability. These self-prediction tasks are sub-optimal solutions, because they are not specifically designed to eliminate the domain bias in the dataset. Contrastive-based methods~\cite{kim2021cds,zhang2021domain} explicitly eliminate the domain bias by pulling the sample and its nearest neighbor from a different domain close. However, the positive pair retrieved by the nearest neighbor across the domains is much more noisy than that in a single domain, because semantically similar images from different domains may have a large visual difference.

In this paper, we tackle the self-supervised learning from multi-domain data from a different perspective, \emph{i.e.,} generative self-supervised learning, and propose a new \textbf{D}omain \textbf{i}nvariant \textbf{M}asked \textbf{A}uto\textbf{E}ncoders (\textbf{DiMAE}) for learning domain-invariant features from multi-domain data, which is motivated by the recent generative-based self-supervised learning method Masked Auto-Encoders (MAE)~\cite{he2021masked}. Specifically, MAE eliminates the low-level information by masking large portion of image patches and drives the encoder to extract semantic information by reconstructing pixels from very few neighboring patches~\cite{cao2022understand} with a light-weighted decoder. However, this design does not take the domain gaps into consideration and thus can not generalize well for the self-supervised learning from multi-domain tasks.
To close the gap, our proposed DiMAE constructs a cross-domain reconstruction task, which uses \emph{the image with the mixed style from different domains
as input for one content encoder to extract domain invariant features and multiple domain-specific decoders to recover the specific domain style for regressing the raw pixel values of masked patches before style mix under an MSE loss}, as shown in Fig.~\ref{fig:di-mae}. The critical designs and insights behind DiMAE for self-supervised learning from multi-domain data involve:

(1) The \textbf{cross-domain reconstruction task} aims at reconstructing the image from the image with other domain styles. DiMAE disentangles the reconstruction into two processes: a content\footnote{``content'' and ``style'' are terms widely used in style mix. ``content'' means domain-invariant information, while ``style'' means domain-specific information.} encoder to remove the domain style by extracting domain-invariant features, and a domain-specific decoder to recover the style of the reconstruction target domain. By forcing the decoder to learn specific style information, we regularize the encoder to learn domain-invariant features.

(2) The \textbf{content preserved style mixing} aims to add style noise of the other domains to one image while preserving the content information. While there exist some popular mixing methods (\emph{e.g,} mixup~\cite{zhang2018mixup} and cutmix~\cite{yun2019cutmix}) able to mix domain styles, they also add content noise to the image. Our experiments find that the content noise will lead to a significant performance decrease in our cross-domain reconstruction task. Therefore, we propose a new non-parametric content preserved style mixing method to take advantage of the cross-domain reconstruction and avoid the undesirable performance decrease by content noise.

(3) The \textbf{multiple domain-specific decoders} aim to recover the corresponding domain style of the target image for reconstruction from the encoded domain-irrelevant features. Although the decoder network design, \emph{e.g.,} such as the number of layers, can determine the semantic level of the learned latent representations as pointed out in MAE~\cite{he2021masked}, we find that a single decoder as used in MAE can not help to regularize the encoder to learn domain-invariant features. To reconstruct the image from a specific domain, the encoder will leak the domain information to guide the decoder to reconstruct the image with the input image's style. This prevents the encoder from learning the domain-invariant features.

Therefore, multiple domain-specific decoders are proposed to recover different domain styles by domain-corresponding decoders, which regularizes the encoder to only learn domain-invariant features.  

To demonstrate the effectiveness of DiMAE, we conduct experiments on the
multi-domain dataset PACS~\cite{li2017deeper} and DomainNet~\cite{peng2019moment}, observing consistent performance improvements on both in-domain and cross-domain settings. For the in-domain evaluation, DiMAE outperforms state-of-the-art methods by \textbf{+0.8\%} on the PACS. On cross-domain testing, we achieve considerable gains over the recent state-of-the-art methods in both linear evaluation and full network fine-tuning. Specifically, in linear evaluation, our method improves the recent state-of-the-art by \textbf{+8.07\%} on PACS with 1\% data fine-tuning fraction. In full network fine-tuning with 100\% data, we get an averaged \textbf{+13.24\%} and \textbf{+9.87\% } performance gains on PACS and DomainNet, respectively.

The contributions of our work are summarized as three-folds: \textbf{(1)} We propose a new generative framework which leverages the cross-domain reconstruction as the pretext to learn domain-invariant features from multi-domain data. \textbf{(2)} We propose a new non-parametric style-mix method that can preserve the content information to exploit the cross-domain reconstruction task and avoid performance drop by content noise. 
\textbf{(3)} We modify the single decoder in MAE to multiple domain-specific decoders to regularize the encoder to learn domain-invariant features. We show that our DiMAE outperforms state-of-the-art self-supervised learning baselines on learning representation from multi-domain data.

\section{Related Work}
\subsection{Self-supervised Learning}
Self-supervised Learning (SSL) introduces various pretext tasks to learn semantic representations from unlabeled data for a better generalization in downstream tasks. Generally, SSL can be categorized into discriminative~\cite{noroozi2016unsupervised,gidaris2018unsupervised,chen2020simple,grill2020bootstrap,he2020momentum,chen2020improved,chen2021empirical,zbontar2021barlow,caron2021emerging} and generative methods~\cite{pathak2016context,larsson2016learning,larsson2017colorization,he2021masked}. Among the former, some early works try to design auxiliary handcrafted prediction tasks to learn semantic representation, such as jigsaw puzzle~\cite{noroozi2016unsupervised} and rotation prediction~\cite{gidaris2018unsupervised}. Recently, contrastive  approaches~\cite{chen2020simple,grill2020bootstrap,he2020momentum,chen2020improved,chen2021empirical,zbontar2021barlow,caron2021emerging} emerge as a promising direction for SSL. They consider each instance a different class and promote the instance discrimination by forcing representation of different views of the same image closer and spreading representation of views from different images apart.

Although remarkable progress has been achieved, contrastive methods heavily rely on data augmentation~\cite{chen2020simple,tian2020makes} and negative sampling~\cite{wu2018unsupervised,he2020momentum}.
Another recent resurgent line of SSL is generative approaches, many of which train an encoder and decoder for pixel reconstruction. Various pretext tasks have been proposed, such as image inpainting~\cite{pathak2016context} and colorization~\cite{larsson2016learning,larsson2017colorization}. Very recently, since the introduction of ViT~\cite{dosovitskiy2020image}, masked image modeling (MIM) has re-attracted the attention of the community. iGPT~\cite{chen2020generative} proposes to predict the next pixels of a sequence, and BEiT~\cite{bao2021beit} leverages a variational autoencoder (VAE) to encode masked patches. A very relevant work, MAE~\cite{he2021masked} 
proposes to train the autoencoder to capture the semantic representation by recovering the input image from very few neighboring patches.
Unlike aforementioned methods that focus on the progress of learning from single domain, our proposed method, a novel generative approach for SSL, is devoted into a more common scenario, pretraining from multiple domains.
As far as we know, we are the first to propose the generative pretraining method for training from multi domain data.

\subsection{Domain Generalization}
Domain Generalization (DG) considers the transferability to unseen target domains using labeled data from a single or multiple source domains. A common approach is to minimize the distance between source domains for learning domain-invariant representations, among which are minimizing the KL Divergence~\cite{wang2021respecting}, minimizing maximum mean discrepancy~\cite{li2018domain} and adversarial learning~\cite{li2018deep,rahman2020correlation,albuquerque2019generalizing}. Several approaches propose to exploit meta-learning~\cite{li2019episodic} or augmentation~\cite{carlucci2019domain,zhou2020deep} to promote the transferability for DG. 

Despite the promising advances in recent DG methods, they assume that source domains are annotated. To address this issue, Unsupervised DG (UDG) is proposed as a more general task of training with unlabeled source domains.~\cite{feng2019self} introduces rotation prediction and mutual information maximization for multi-domain generalization. Derived from contrastive learning, DIUL~\cite{zhang2021domain} incorporates domain information into the contrastive loss by a reweighting mechanism considering domain labels. Despite the promising results, these two works carefully design domain-related discriminative pretext tasks and try to strike a compromise between instance and domain discrimination.
Our proposed method, in contrast, is a brand new generative approach for self-supervised learning from multi-domain data, showing strong advantages for UDG setting.

\begin{figure*}[t]
    \centering
    \includegraphics[width=0.9\linewidth]{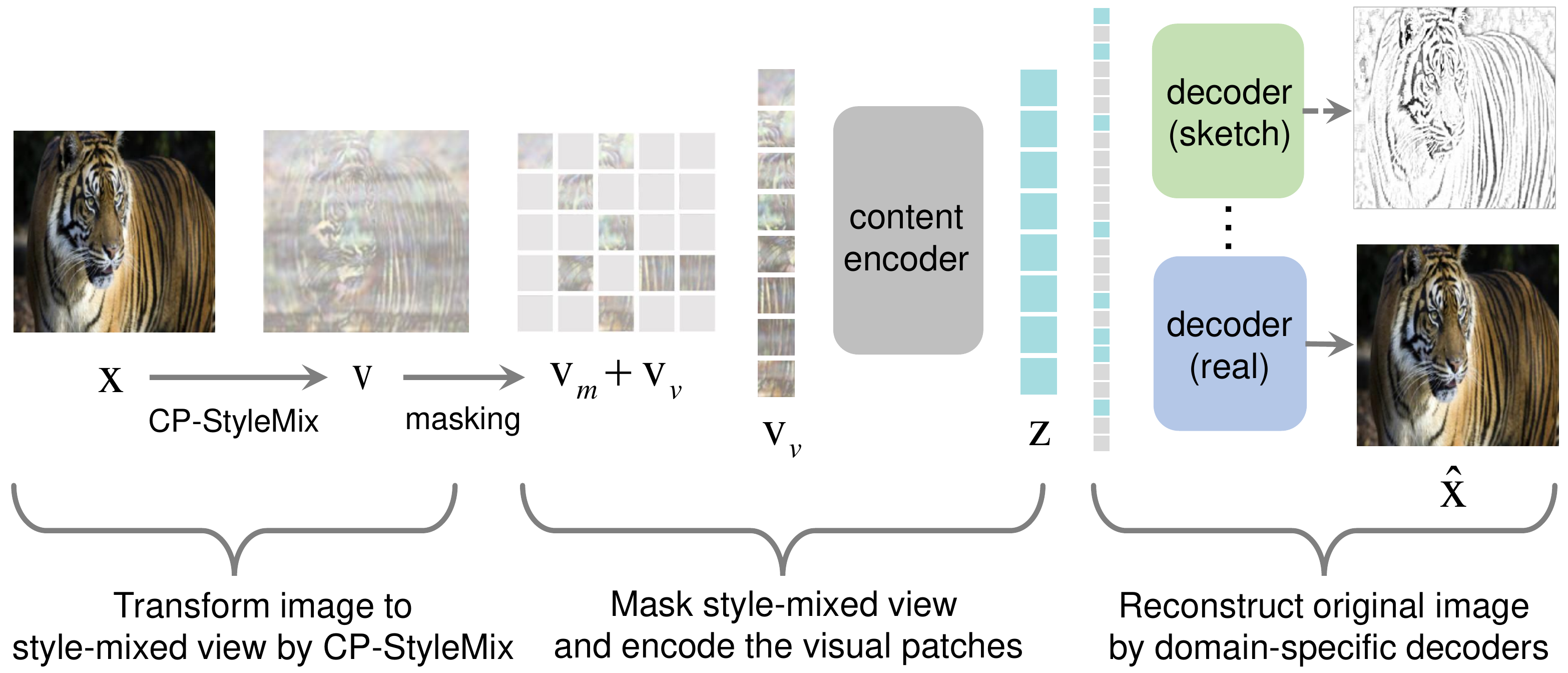}
    \caption{The pipeline of DiMAE. First, CP-StyleMix transforms the original image $\mathbf{x}$ to its style-mixed view $\mathbf{v}$ by adding style information from other domains without introducing content noise. Second, the style-mixed view $\mathbf{v}$ is divided into visible patches $\mathbf{v}_v$ and masked patches $\mathbf{v}_m$, and the content encoder learns the content representation $\mathbf{z}$ from visible patches. Third, domain-specific decoders learn to reconstruct $\hat{\mathbf{x}}$ by the corresponding decoder.}
    
    \label{fig:di-mae}
\end{figure*}

\section{Domain-Invariant Masked AutoEncoder}
\subsection{Cross-domain Reconstruction Framework}
Different from MAE which learns high-level semantic representations by reconstruction from a highly masked image, our DiMAE learns domain invariant representation by a cross-domain reconstruction task, which aims at recovering images from an image mixed with other domain styles. Specifically, DiMAE consists three modules, including a Content Preserved Style-Mix (CP-StyleMix), a content encoder, and multiple domain-specific decoders. The CP-StyleMix is used to mix the style information from different domains while preserving the domain-irrelevant object content, which generates the input of the cross-domain reconstruction task. The content encoder $\mathcal{F}(*, \theta_{\mathcal{F}})$ are shared by images from all domains, where $\theta_{\mathcal{F}}$ is the parameter of $\mathcal{F}$, and is expected to encode the content and domain-invariant information by denoising the style information. The domain-specific decoders $\mathcal{G}$ in DiMAE are designed to incorporate the style information to the domain-invariant representation for image reconstruction, where $\mathcal{G}=\{\mathcal{G}_1(*, \phi_1), \mathcal{G}_2(*, \phi_2), ..., \mathcal{G}_{N_d}(*, \phi_{N_d})\}$, $\phi_{i}$ is the parameter for the $i$-th domain-specific decoder and $N_d$ is the number of domains in the training set. As shown in Fig.~\ref{fig:di-mae}, our DiMAE has the following steps:

\emph{Step1}: \emph{Transform an image $\mathbf{x}$ to its style-mixed view $\mathbf{v}$ by Content Preserved Style-Mix (Sec.~\ref{sec:Content Preserved Style-Mix}).} Given an image $\textbf{x}$, with Content Preserved Style-Mix, we mix the style from other domains to the image $\mathbf{x}$ while preserving the content in $\mathbf{x}$ to generate its style-mixed view $\mathbf{v}$.

\emph{Step2}: \emph{Transform the style-mixed view $\mathbf{v}$ to content representation $\mathbf{z}$ (Sec.~\ref{sec:Content Encoder}).} We randomly divides $\mathbf{v}$ into visible patches $\mathbf{v}_v$ and masked patches $\mathbf{v}_m$, and extract content representation $\mathbf{z}$ by encoding the visible patches $\mathbf{v}_v$ by $\mathcal{F}(*, \theta_{\mathcal{F}})$. 

\emph{Step3}: \emph{Reconstruct the image $\hat{\mathbf{x}}$ by content representation $\mathbf{z}$ with the domain-specific decoders (Sec.~\ref{sec:Domain Specific Decoder}).} Given content representation $\textbf{z}$ and multiple domain-specific decoders $\mathcal{G}=\{\mathcal{G}_1(*, \phi_1), \mathcal{G}_2(*, \phi_2), ..., \mathcal{G}_{N_d}(*, \phi_{N_d})\}$, we reconstruct the image $\hat{\mathbf{x}}$ by $\mathcal{G}_i$, where $\mathcal{G}_i$ is the decoder of the $i$-th domain. 

\emph{Step4}: \emph{Backward propagation using the MSE loss (Sec.~\ref{sec:Objective}).} Given the reconstructed image $\hat{\mathbf{x}}$ and the original image $\mathbf{x}$, the parameters $\theta_{\mathcal{F}}$ in $\mathcal{F}(*, \theta_{\theta_{\mathcal{F}}})$ and the parameters $\phi_1, \phi_2, \cdots, \phi_{N_d}$ in $\mathcal{G}(*, \phi_1)$, $\mathcal{G}(*, \phi_2)$, ..., $\mathcal{G}(*, \phi_{N_d})$ are learned by the MSE loss.

\begin{figure}[t]
    \centering
    \includegraphics[width=0.95\linewidth]{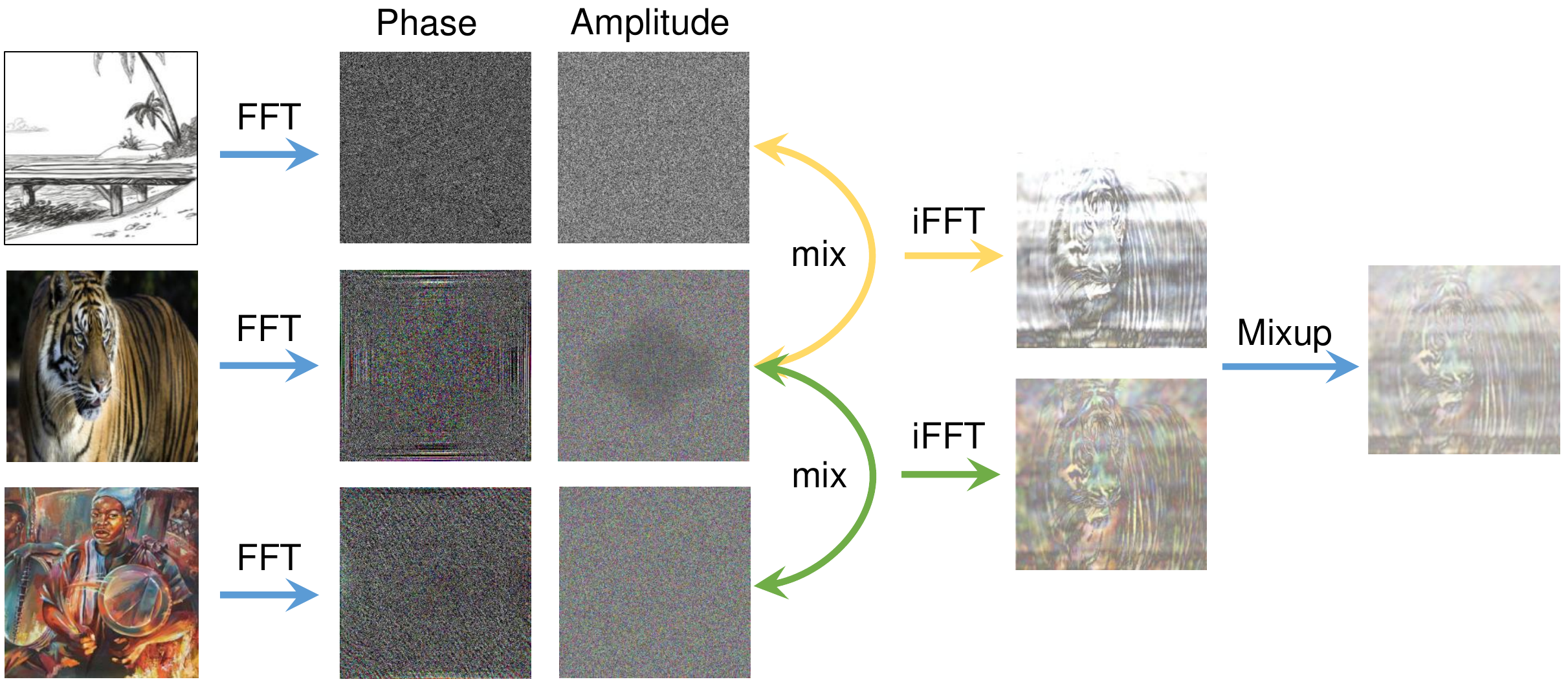}
    \caption{The pipeline of CP-StyleMix. We mix the Fourier Amplitude of the original image $\textbf{x}$ and two images from other domains to generate content preserved and style-transferred images, and mix them to generate the style-mixed view \textbf{v}. 
    }
    \label{fig:cpsm}
\end{figure}

\subsection{Content Preserved Style-Mix}
\label{sec:Content Preserved Style-Mix}
Content Preserved Style-Mix (CP-StyleMix) aims at mixing two styles into an image while preserving the content information. This is a critical part for the cross-domain reconstruction tasks. Inspired by~\cite{xu2021fourier}, the style information and the content information can be disentangled in the Fourier space. The content information is encoded in the phase of Fourier signals, and the style information is encoded in the amplitude of the Fourier signals. We propose to first mix the style of the $i$-th domain to the image $\mathbf{x}$ in the Fourier space, generation its style views $\{\mathbf{v}_1, \mathbf{v}_2, ..., \mathbf{v}_{N_d}\}$, where $N_d$ is the number of domains. Then we mix these style views by the typical Mixup method~\cite{yun2019cutmix} in the image space, generating the final style-mixed view $\mathbf{v}$.

Specifically, for mixing in the Fourier space, given an image $\mathbf{x}$ from $j$-th domain and a randomly selected image $\mathbf{x}_{aux}$ from the $i$-th domain ($i\ne j$), the view $\mathbf{v}_i$ of image $\mathbf{x}$ can be formulated as
\begin{equation}
    \mathbf{v}_i = \mathcal{K}^{-1}(\mathcal{K}^{A}_{mix}, \mathcal{K}^{P}(\mathbf{x})),
    \label{eq:am-mix}
\end{equation}
where $\mathcal{K}^{A}_{mix} = \lambda \mathcal{K}^{A}(\mathbf{x}_{aux}) + (1-\lambda)\mathcal{K}^{A}(\mathbf{x})$, $\mathcal{K}^{-1}$ is Fourier inversion, and $\mathcal{K}^{A}$, $\mathcal{K}^{P}$ returns the amplitude and phase of Fourier transformation, respectively. Then we implement the second step of mix on the image space by Mixup~\cite{zhang2018mixup} process. Mathematically, the Mixup process can be formulated as
\begin{equation} \label{eq:mixup}
    \mathbf{v} = \sum_{i=1}^{N_d} \mu_i  \mathbf{v}_i,
\end{equation}
where $\mu_i$ is the weight of different views, $\sum_{i=1}^{N_d} \mu_i=1$, $\mu_j=0$. Different from the Fourier style transfer proposed by~\cite{xu2021fourier}, which do not have style mix, we mix different styles in both Fourier space and image space, leading to more diverse style information.

\noindent \textbf{Discussion.} 
Theoretically, as summarized in Tab.~\ref{tab:augmentations}, there are various methods to mix the style information from other domains to the input image, including CutMix~\cite{yun2019cutmix}, MixUp~\cite{zhang2018mixup}, StyleMix~\cite{hong2021stylemix}, and CycleGan+Mix\cite{zhu2017unpaired}. Our content preserved style mix is better than these methods in two critical aspects. First, our CP-StyleMix can preserve content information compared to CutMix and Mixup, which also mix contents. Detailed experiments and analysis in Sec.~\ref{sec:ablation} illustrates that compared with content-pereserved methods, the mixture of content with Mixup and CutMix would significantly decrease the performance in reconstruction tasks by $-10.47\%$ and $-9.71\%$, respectively. Second, our CP-StyleMix is non-parametric and does not need extra data. StyleMix~\cite{hong2021stylemix} and CycleGan+Mix\cite{zhu2017unpaired} can preserve the content information, but they require to train the transfer module by extra data, which will lead to unfair comparison with existing methods~\cite{feng2019self,zhang2021domain}.

\begin{table*}[t]
  \small
  \centering
  \caption{Comparison between existing augmentation methods and CP-StyleMix. All these existing methods do not fully meet the requirements of being both content preserved and light-weighted. }
    \begin{tabular}{m{35mm}m{20mm}<{\centering}m{20mm}<{\centering}m{20mm}<{\centering}}
    \toprule
        Method & Venue & Content preserved & No extra training \\
    \hline
    CutMix~\cite{yun2019cutmix} & ICCV'2019   &    & \checkmark \\
    MixUp~\cite{zhang2018mixup} &  ICLR'2018  &   &  \checkmark\\
    StyleMix~\cite{hong2021stylemix} & CVPR'2021 & \checkmark    &  \\
    StyleCutMix~\cite{hong2021stylemix} & CVPR'2021 &  \checkmark   &  \\
    CycleGan+Mix~\cite{zhu2017unpaired}  & ICCV'2017 & \checkmark     &  \\
    CP-StyleMix(ours) & - & \checkmark     & \checkmark \\
    \bottomrule
    \end{tabular}%
  \label{tab:augmentations}%
\end{table*}%

\subsection{Content Encoder}
\label{sec:Content Encoder}
The content Encoder, \emph{i.e.,} $\mathcal{F}(*, \theta_{\mathcal{F}})$, is designed to extract the domain-invariant content representations from the style-mixed view $\mathbf{v}$. Similar to MAE~\cite{he2021masked}, our content encoder also follows the vision transformer design, which extracts content representations only by visible patches. Specifically, given a style-mixed view $\mathbf{v}$, we randomly divide the image patches into visible patches $\mathbf{v}_v$ with the probability $p$, leaving the remaining patches as the masked patches $\mathbf{v}_m$. The content representation $\mathbf{z}$ is then extracted by $\mathbf{v}_v$ using the content encoder, \emph{i.e,}
\begin{equation}
    \mathbf{z} = \mathcal{F}(\mathbf{v}_v, \theta_{\mathcal{F}}).
\end{equation}

\subsection{Domain Specific Decoders}
\label{sec:Domain Specific Decoder}
Domain specific decoders are the critical designs in our proposed DiMAE. Besides the target of the decoder in MAE that is to reconstruct the semantic meaning of the masked patches, Domain specific decoders are expected to additionally reconstruct the domain style of the masked patches. To achieve this, we design a domain-specific decoder to each domain in the training set. Specifically, the domain specific decoders are defined as $\mathcal{G}=\{\mathcal{G}_1(*, \phi_1), \mathcal{G}_2(*, \phi_2), ..., \mathcal{G}_{N_d}(*, \phi_{N_d})\}$, where $N_d$ is the number domains in the training set, $\mathcal{G}_1, \mathcal{G}_2, ..., \mathcal{G}_{N_d}$ share the same architectural design, and $\phi_{i}$ is the parameter of the $i$-th domain-specific decoder $\mathcal{G}_i$. Given content representation $\mathbf{z}$, to reconstruct the patches in the $i$-th domain, we feed both the content representation $\mathbf{z}$ and the learnable masked tokens~\cite{he2021masked} into the $i$-th domain specific decoder $\mathcal{G}_i$, \emph{i.e.,}
\begin{equation}
    \hat{\mathbf{v}}_m^i = \mathcal{G}_i(\mathbf{z}, \mathbf{q}_m^i),
\end{equation}
where $i\in [1, N_d]$ denotes the domain index, and the $\mathbf{q}_m^i$ denotes the masked tokens in the $i$-th domain-specific decoder.

\noindent \textbf{Discussion.} As pointed in MAE~\cite{he2021masked}, the decoder design plays a key role in determining the semantic level of the learnt latent features. However, we argue that the domain-invariant features can not be learnt by changing the single decoder designs probably because of the style conflict in different domains. Instead, we propose to use multiple domain-specific decoders to learn the domain-invariant features. Specifically, we use a shared content encoder to learn the domain-invariant features, and expect the domain-specific decoder to recover the specific style information for the cross-domain reconstruction.

\subsection{Objective Function}
\label{sec:Objective}
The objective function constrains the error between predicted patches and target patches, which drives the model to recover the original image $\mathbf{x}$ using very few mixed-styled neighboring patches. Specifically, given the image $\mathbf{x}$ from the $j$-th domain, the objective function can be formulated as
\begin{equation}
    \mathcal{L} = (\mathbf{\hat{v}}^j_m - \mathbf{x}_m)^2,
    \label{eq:object function}
\end{equation}
where $\mathbf{\hat{v}}^j_m$ is the reconstructed masked patch by $\mathcal{G}_j$, $\mathbf{x}_m$ is the corresponding masked patches in the original image $\mathbf{x}$.

\section{Experiment}
\subsection{Experimental Setup}
\noindent \textbf{Dataset.}
To validate our approach, we conduct extensive experiments with two generalization settings, namely in-domain and cross-domain, which detailed in Sec. \ref{Experimental Results}. Two benchmark datasets are adopted to carry through these two settings. PACS~\cite{li2017deeper} is a widely used benchmark for domain generalization. It consists of four domains, including Photo (1,670 images), Art Painting (2,048 images), Cartoon (2,344 images) and Sketch (3,929 images) and each domain contains seven categories. DomainNet~\cite{peng2019moment} is the largest, most diverse and recent cross-domain benchmark. Six domains are included: Real, Painting, Sketch, Clipart, Infograph and Quickdraw, with 345 object classes and 586, 575 examples. 

\underline{For In-domain evaluations}, we use all training subset in all domains for self-supervised learning, and then use the validation subset of each domain for evaluation. \underline{For cross-domain generalization}, following DIUL~\cite{zhang2021domain}, we select Painting, Real, Sketch as source domains and Clipart, Infograph, Quickdraw as target domains for DomainNet~\cite{peng2019moment}. We select 20 classes out of 345 categories for both training and testing, exactly following the setting in~\cite{zhang2021domain}. For PACS, we follow the common setting in domain generalization~\cite{li2018deep,rahman2020correlation,albuquerque2019generalizing} where three domains are selected for self-supervised training, and the remaining domain is used for evaluation.

\noindent \textbf{Implementation details.}
In our implementation, we use ViT-small~\footnote{We do not use the widely-used ResNet18~\cite{he2016deep} as the backbone, because DiMAE is exactly a generative method, in which Convolutaional networks are not applicable. We choose the ViT-small model for comparison because the number of their model parameters is similar. } as the backbone network unless otherwise specified. The learning rate for pretraining is $1.5 \times 10^{-4}$ and then decays with a cosine
decay schedule. The weight decay is set to 0.05 and the batch size is set to $256 \times N_d$, where $N_d$ is the number of domains in the training set. All methods are pretrained for 1000 epochs, which is consistent with the implementations in~\cite{zhang2021domain} for fair comparison. The feature dimension is set to 1024. For finetuning, we follow the exact training schedule as that in~\cite{zhang2021domain}. Following~\cite{kim2021cds}, we use an ImageNet pretraining.

\begin{table*}[t]
    \small
    \centering
    \caption{Results of In-domain top-1 linear evaluation accuracies on PACS dataset. Results style: \textbf{best}, \underline{second best}.}
    \begin{tabular}{p{30mm}<{\centering}|p{15mm}<{\centering}p{15mm}<{\centering}p{15mm}<{\centering}p{15mm}<{\centering}|p{15mm}<{\centering}}
        \hline
         Training Domain & \multicolumn{4}{c|}{(Photo, Art, Cartoon, Sketch)} \\
         \hline
         Method & Photo & Art & Cartoon & Sketch & Avg. \\
        \hline
        MoCo V3 & 70.6 &39.4 &64.8 & 54.4 & 57.3 \\
        MAE & 83.5	&53.4&	\underline{74.2} & \textbf{73.8} & \underline{71.2} \\
        DeepAll+MI,RotNet & \underline{81.6} & 55.5 & 68.5 & 63.4 & 67.3  \\
        DeepAll+MI,AET & 80.9 & \underline{56.9} & 69.6 & 67.9 & 68.8  \\
        \hline
        DiMAE (ours) & \textbf{84.7} & \textbf{57.2} & \textbf{76.3} & \underline{69.8} & \textbf{72.0}  \\
        \hline
    \end{tabular}
    \label{tab-In-Domain-PACS}
\end{table*}

\begin{table*}[t]
    \centering
    \caption{Results of the cross-domain generalization on DomainNet. All of the models are trained on Painting, Real, Sketch domains of DomainNet and tested on the other three domains. The title of each column indicates the name of the domain used as target. All the models are pretrained for 1000 epoches before finetuned on the labeled data. Results style: \textbf{best}, \underline{second best}.}
    \resizebox{1\textwidth}{!}{
    \begin{tabular}{c|ccc|cc|ccc|cc}
        \toprule
         & \multicolumn{5}{c|}{Label Fraction 1\%} & \multicolumn{5}{c}{Label Fraction 5\%} \\
         \hline
         method & Clipart & Infograph & Quickdraw & Overall & Avg. & Clipart & Infograph & Quickdraw & Overall & Avg. \\
        \hline
        ERM  & 6.54 & 2.96 & 5.00 & 4.75 & 4.83 & 10.21 & 7.08 & 5.34 & 6.81 & 7.54 \\
        MoCo V2 \cite{chen2020improved}  & 18.85 & 10.57 & 6.32 & 10.05  & 11.92 & 28.13 & 13.79 & 9.67 & 14.56 & 17.20 \\
        SimCLR V2 \cite{chen2020big}  & \underline{23.51} & \underline{15.42} & 5.29 & 11.80 & 14.74 & 34.03 & 17.17 & 10.88 & 17.32 & 20.69 \\
        BYOL \cite{grill2020bootstrap} & 6.21 & 3.48 & 4.27 & 4.45 & 4.65 & 9.60 & 5.09 & 6.02 & 6.49 & 6.90 \\
        AdCo \cite{hu2021adco}  & 16.16 & 12.26 & 5.65 & 9.57 & 11.36 & 30.77 & 18.65 & 7.75 & 15.44 & 19.06 \\
        MAE & 22.38 & 12.62 & 10.50 & \underline{13.51} & \underline{15.17} & 32.60&15.28 &\underline{13.43}& 17.85 & 20.44 \\
        DIUL   & 18.53 & 10.62 & \underline{12.65} & 13.29 & 13.93 & \underline{39.32} & \textbf{19.09} & 10.50 & \underline{18.73} & \underline{22.97} \\
        \hline
        DiMAE (ours)  & \textbf{26.52} & \textbf{15.47} & \textbf{15.47} & \textbf{17.72} & \textbf{19.15} & \textbf{42.31} & \underline{18.87} & \textbf{15.00} & \textbf{21.68} & \textbf{25.39} \\
        \hline
        & \multicolumn{5}{c|}{Label Fraction 10\%} & \multicolumn{5}{c}{Label Fraction 100\%} \\
         \hline
         method & Clipart & Infograph & Quickdraw & Overall & Avg. & Clipart & Infograph & Quickdraw & Overall & Avg. \\
        \hline
        ERM  & 15.10 & 9.39 & 7.11 & 9.36 & 10.53 & 52.79 & 23.72 & 19.05 & 27.19 & 31.85 \\
        MoCo V2 & 32.46 & 18.54 & 8.05 & 15.92 & 19.69 & 64.18 & 27.44 & 25.26 & 33.76 & 38.96 \\
        SimCLR V2 & 37.11 & 19.87 & 12.33 & 19.45 & 23.10 & 68.72 & 27.60 & 30.56 & 37.47 & 42.29 \\
        BYOL & 14.55 & 8.71 & 5.95 & 8.46 & 9.74 & 54.44 & 23.70 & 20.42 & 28.23 & 32.86 \\
        AdCo & 32.25 & 17.96 & 11.56 & 17.53 & 20.59 & 62.84 & 26.69 & 26.26 & 33.80 & 38.60 \\ 
        MAE  & \underline{51.86} & \underline{24.81} & \underline{23.94} & \underline{29.87} & \underline{33.54} & 59.21 & 28.53 & 23.27 & 32.06 & 37.00 \\
        DIUL   & 35.15 & 20.88 & 15.69 & 21.08 & 23.91 & \underline{72.79} & \underline{32.01} & \underline{33.75} & \underline{41.19} & \underline{46.18} \\
        \hline
        DiMAE (ours)  & \textbf{70.78} & \textbf{38.06} & \textbf{27.39} & \textbf{39.20} & \textbf{45.41} & \textbf{83.87} & \textbf{44.99} & \textbf{39.30} & \textbf{49.96} & \textbf{56.05} \\
        \bottomrule
    \end{tabular}}
    \label{tab:all_correlated_domainet}
    \end{table*}

\subsection{Experimental Results}
\label{Experimental Results}
\noindent \textbf{In-Domain Evaluation.}
In-Domain Evaluation is proposed by~\cite{feng2019self}, and aims to evaluate the performance of the self-supervised learning methods in the domains that appear in the training set. We exactly follow the protocol of \cite{feng2019self}. Specifically, we learn the backbone on the training subset of Photo, Art, Cartoon and Sketch on PACS in a self-supervised manner, and then linearly train a classifier for each domain using the training subset of each domain with the backbone fixed, respectively. We evaluate our model on the validation subset in each domain, and report the averaged results by 10 runs. The experimental results are summarized in Tab.~\ref{tab-In-Domain-PACS}.
DiMAE outperforms MoCo V3 and MAE by \textbf{+14.7\%} and \textbf{+0.8\%}, respectively, showing the superior of in-domain instance discrimination ability against the previous methods. Furthermore, when we compare the baseline generative method, \emph{i.e.,} MAE, with contrastive learning methods, \emph{i.e.,} MoCoV3, we infer that the reconstruction task can learn better representations of the domains that appear in the training set.

\noindent \textbf{Cross-Domain Generalization.}
Cross-Domain Generalization is firstly proposed by DIUL~\cite{zhang2021domain}, which evaluates the generalization ability of the self-supervised learning methods to the domains that are missing in the training set. We exactly follow the cross-domain generalization evaluation process in DIUL~\cite{zhang2021domain}, which is divided into three steps. First, we train our model on source domains in the unsupervised manner. Then, we will use a small number of labeled training examples of the validation subset in the source domains to finetune the classifier or the whole backbone. In detail, when the fraction of labeled finetuning data is lower than 10\% of the whole validation subset in the source domains, we only finetune the linear classifier for all the methods. When the fraction of labeled finetuning data is larger than 10\% of the whole validation subset in the source domains, we finetune the whole network, including the backbone and the classifier. Last, we can evaluate the model on the target domains.

The results are presented in Tab.~\ref{tab:all_correlated_domainet} (DomainNet) and Tab.~\ref{tab:all_correlated_pacs} (PACS). In this setting, our DiMAE achieves  a better performance than previous works on most tasks and gets significant gains over DIUL and other SSL methods on overall and average accuracy\footnote{Overall and Avg. indicate the overall accuracy of all the test data and the arithmetic mean of the accuracy of 3 domains, respectively. Note that they are different because the capacities of different domains are not equal.}. Compared with contrastive learning based methods, such as MoCo V2, SimCLR V2, BYOL, AdCo, our generative based methods improves the cross-domain generalization tasks by \textbf{+3.98\%} and \textbf{+2.42\%} for DomainNet and \textbf{+8.07\%} and \textbf{+0.23\%}  for PACS on 1\% and 5\% fraction setting respectively, which is tested by linear evaluation. Our DiMAE also improves other states-of-the-art methods by \textbf{+11.87\%} and \textbf{+9.87\%} for DomainNet, \textbf{+16.18\%} and \textbf{+13.24\%} for PACS on 10\% and 100\% fraction setting, respectively, when the whole backbone are finetuned. The significant improvement to contrastive learning based methods illustrate our proposed DiMAE  can learn more domain-invariant features in the self-supervised learning from multiple domain data.

\begin{table*}[t]
    \centering
    \footnotesize
    \caption{Results of the cross-domain generalization setting on PACS. Given the experiment for each target domain is run respectively, there is no overall accuracy across domains. Thus we report the average accuracy and the accuracy for each domain. The title of each column indicates the name of the domain used as target. All the models are pretrained for 1000 epochs before finetuned on the labeled data. Results style: \textbf{best}, \underline{second best}.}
    \resizebox{0.96\textwidth}{!}{
    \begin{tabular}{c|cccc|c|cccc|c}
        \toprule
         & \multicolumn{5}{c|}{Label Fraction 1\%} & \multicolumn{5}{c}{Label Fraction 5\%} \\
         \hline
         method & Photo & Art. & Cartoon & Sketch & Avg. & Photo & Art. & Cartoon & Sketch & Avg. \\
        \hline
    
        MoCo V2 & 22.97 & 15.58 & 23.65 & 25.27 & 21.87 & 37.39 & 25.57 & 28.11 & 31.16 & 30.56 \\
        SimCLR V2 & \underline{30.94} & 17.43 & \textbf{30.16} & 25.20 & 25.93 & \textbf{54.67} & 35.92 &\underline{35.31} & \underline{36.84} & \underline{40.68} \\
        BYOL & 11.20 & 14.53 & 16.21 & 10.01 & 12.99 & 26.55 & 17.79 & 21.87 & 19.65 & 21.47 \\
        AdCo & 26.13 & 17.11 & 22.96 & 23.37 & 22.39 & 37.65 & 28.21 & 28.52 & 30.35 & 31.18 \\ 
        MAE  & 30.72 & \underline{23.54} & 20.78 & 24.52 & 24.89 & 32.69	&24.61 & 27.35 & 30.44 & 28.77 \\
        
        DIUL   & 27.78 & 19.82 & \underline{27.51} & \underline{29.54} & \underline{26.16} & 44.61 & \underline{39.25} & \textbf{36.41} & 36.53 & 39.20 \\
        \hline
        DiMAE (ours)  & \textbf{48.86} & \textbf{31.73} & 25.83 & \textbf{32.50} & \textbf{34.23} & \underline{50.00} & \textbf{41.25} & 34.40 & \textbf{38.00} & \textbf{40.91} \\
        \hline
        & \multicolumn{5}{c|}{Label Fraction 10\%} & \multicolumn{5}{c}{Label Fraction 100\%} \\
         \hline
         method & Photo & Art. & Cartoon & Sketch & Avg. & Photo & Art. & Cartoon & Sketch & Avg. \\
        \hline
    
        MoCo V2 & 44.19 & 25.85 & 33.53 & 24.97 & 32.14 & 59.86 & 28.58 & 48.89 & 34.79 & 43.03 \\
        SimCLR V2 & \underline{54.65} & 37.65 & 46.00 & 28.25 & 41.64 & 67.45 & \underline{43.60} & 54.48 & 34.73 & 50.06\\
        BYOL & 27.01 & 25.94 & 20.98 & 19.69 & 23.40 & 41.42 & 23.73 & 30.02 & 18.78 & 28.49 \\
        AdCo & 46.51 & 30.21 & 31.45 & 22.96 & 32.78 & 58.59 & 29.81 & 50.19 & 30.45 & 42.26 \\  
        MAE  & 35.89 & 25.59 & 33.28 & \underline{32.39} & 31.79 & 36.84 & 25.24	&32.25 & 34.45 & 32.20 \\
        DIUL  & 53.37 & \underline{39.91} & \underline{46.41} & 30.17 & \underline{42.47} & \underline{68.66} & 41.53 & \underline{56.89} & \underline{37.51} & \underline{51.15} \\
        \hline
        DiMAE (ours)  & \textbf{77.87} & \textbf{59.77} &  \textbf{57.72} & \textbf{39.25}  & \textbf{58.65}  & \textbf{78.99} & \textbf{63.23} & \textbf{59.44} & \textbf{55.89} & \textbf{64.39} \\
        \bottomrule
    \end{tabular}}
    \label{tab:all_correlated_pacs}

\end{table*}

\subsection{Ablation Study}
\label{sec:ablation}
To investigate the effectiveness of each component of our proposed DiMAE, We ablate our DiMAE on the Cross-Domain Generalization task. Specifically, we train Vit-Tiny~\cite{touvron2021training} for 100 epoches on the combination of Painting, Real, and Sketch training set in DomainNet, and evaluate the model using the linear evaluation protocol on Clipart.

\noindent \textbf{Effectiveness of Preserving Contents in Style Mix.}
To demonstrate the importance of preserving contents in style mix, we ablate the content-preserved and content-mix augmentation methods for DiMAE, which is presented in Tab.~\ref{tab_content_mix}. Specifically, we choose CP-StyleMix for content-preserved methods and Mixup and CutMix for content-mixed methods.
Additionally, to fairly compare with CutMix, we replace the Mixup step in Content Preserved StyleMix with CutMix, creating a competing method called Content Preserved StyleCut (CP-StyleCut). 
The experimental results of these methods are illustrated in Tab.~\ref{tab_content_mix}. We conclude that preserving the content information is critical for reconstruction tasks. Specifically, we observe that content-mix methods, \emph{i.e.,} Mixup and CutMix, bring at most $+\textbf{1.24}\%$ performance improvement compared with no augmentation. However, two content preserved style mix methods, \emph{i.e,} CP-StyleMixp and CP-StyleCut, can further improve the content-mix style-mix augmentations, \emph{i.e.,} Mixup and CutMix, by $+\textbf{10.47}\%$ and $+\textbf{9.71}\%$. The large performance gap between content-preserved and content-mix augmentations methods indicates the importance of preserving contents in the reconstruction tasks. 

\begin{table*}[t]
    \centering
\begin{tabular}{p{20mm}<{\centering}p{20mm}<{\centering}|p{20mm}<{\centering}p{20mm}<{\centering}|p{20mm}<{\centering}}
    \hline
    \multicolumn{2}{c|}{Content-preserved} & \multicolumn{2}{c|}{Content-mix} & \multirow{1}[2]{*}{No aug.} \\
 \cline{1-4}     
\multicolumn{1}{c}{CP-StyleMix} & \multicolumn{1}{c|}{CP-StyleCut} & \multirow{1}{*}{Mixup} & \multirow{1}{*}{CutMix} &  \\
    \hline
    \textbf{48.56} & 47.21 & 38.09 & 37.50  & 36.85 \\
    \hline
    \end{tabular}%
    \caption{Comparison of using content-preserved methods, content-mix methods, and no augmentation. Aug. is short for augmentation.}
    \label{tab_content_mix}
\end{table*}

\noindent \textbf{Effectiveness of Mixing Style Information.} To illustrate the importance of mixing style information in our propose DiMAE (Eq.~\ref{eq:mixup}), we ablate the mixing step by comparing the experiments where we use the mixed-style view $\mathbf{v}$ in Eq.~\ref{eq:am-mix}, and the view $\mathbf{v}_i$ before mixing. Here, $\mathbf{v}_i$ is the $i$-th style view after style transfer (Eq.~\ref{eq:am-mix}) before Mixup (Eq.~\ref{eq:mixup}). As shown in Tab.~\ref{tab_style_mix_and_style_transfer}, after applying Mixup and CutMix on the view after style transfer, the performance of the model further increases by $+\textbf{2.45}\%$ to $+\textbf{1.10}\%$, respectively. The consistent improvement indicates that adding more style noise by style mixing can effectively help the encoder to learn domain-invariant features. 

\begin{table}[t]
    \centering
     \begin{tabular}{m{35mm}<{\centering}|m{10mm}<{\centering}}
     \hline
    \multicolumn{1}{c|}{Content-preserved Augmentation} & \multirow{1}{*}{Top-1} \\
    \hline
    Style transfer~\cite{xu2021fourier}   & 46.11 \\
    CP-StyleMix & \textbf{48.56} \\
    CP-StyleCut & 47.21 \\
    \hline
    \end{tabular}%
    \caption{Comparison of style transfer~\cite{xu2021fourier}, CP-StyleMix and CP-StyleCut. 
Aug. is short for augmentation.}
    \label{tab_style_mix_and_style_transfer}
\end{table}

\noindent \textbf{Effectiveness of Multiple Domain-specific Decoders.} A novel design of our proposed DiMAE is the domain-specific decoders, which reconstruct corresponding domain-specific images using the encoded latent representation. We ablate this design with all other factors fixed. Experimental results are illustrated in Tab.~\ref{tab-Domain-Specific-Decoder}, showing the linear evaluation performance when the single decoder and Domain Specific Decoders are applied. We observe that the methods using domain-specific decoders improve the methods using the single decoder by $+\textbf{10.47}\%$ and $+\textbf{9.71}\%$ when images are augmented by CP-StyleMix and CP-StyleCut, respectively. The significant performance gap between two methods verifies the importance of using multiple domain-specific decodoers in our proposed DiMAE. To explain the performance gap, we argue that this is because domain-specific decoders help to decouple the different style information from different domains to the corresponding decoders, regularizing the encoder to only learn domain-invariant features.

\begin{table}[t]
    \centering
    \footnotesize
    \begin{tabular}{m{20mm}<{\centering}|m{20mm}<{\centering}|m{20mm}<{\centering}}
    \hline
          Augmentations & Single Decoder & Domain Specific Decoders \\
          \hline
    {CP-StyleMix} & 38.09 & \textbf{48.56} \\
          {CP-StyleCut} & 37.50  & \textbf{47.21} \\
          \hline
    \end{tabular}
    \caption{Comparison of single decoder and Domain Specific Decoders. 
Domain Specific Decoders achieve significant performance improvement 
with CP-StyelMix and CP-StyleCut.}
    \label{tab-Domain-Specific-Decoder}
\end{table}

\begin{table}[t]
    \centering
    \footnotesize
    \begin{tabular}{m{20mm}<{\centering}|m{20mm}<{\centering}|m{20mm}<{\centering}}
    \hline
    Depth & Single Decoder & Multi Decoders \\
    \hline
    1     & 37.46 & 44.93 \\
    2     & 37.81 & 45.35 \\
    4     & 38.01 & 46.62 \\
    8     & \textbf{38.09} & \textbf{48.56} \\
    12    & 37.96 & 46.11 \\
    \hline
    \end{tabular}%
    \caption{Comparison of different depth of Domain Specific Decoders.}
    \vspace{-1em}
    \label{tab-Decoder_design}
\end{table}

\noindent \textbf{Designs in the single decoder and multiple domain-specific decoders.}
Tab.~\ref{tab-Decoder_design} varies the decoder depth (number of Transformer blocks), from which we have two findings. First, we find the depth of the decoder is also important in our task, because a sufficiently deep decoder can improves the performance by 0.63\% and 3.63\% in single and multiple decoders design, respectively. Second, the performance gain in multi-decoders design (\textbf{+3.63\%}) is much larger than in single-decoder design (\textbf{+0.63\%}), because the depth of decoders can influence the semantic level of the learned feature, but can not help to regularize the encoder to learn domain-invariant features, which is crucial in our self-supervised learning from multi-domain data task.

\subsection{Visualization}

\begin{figure}[t]
    \centering
    \includegraphics[width=.95\linewidth]{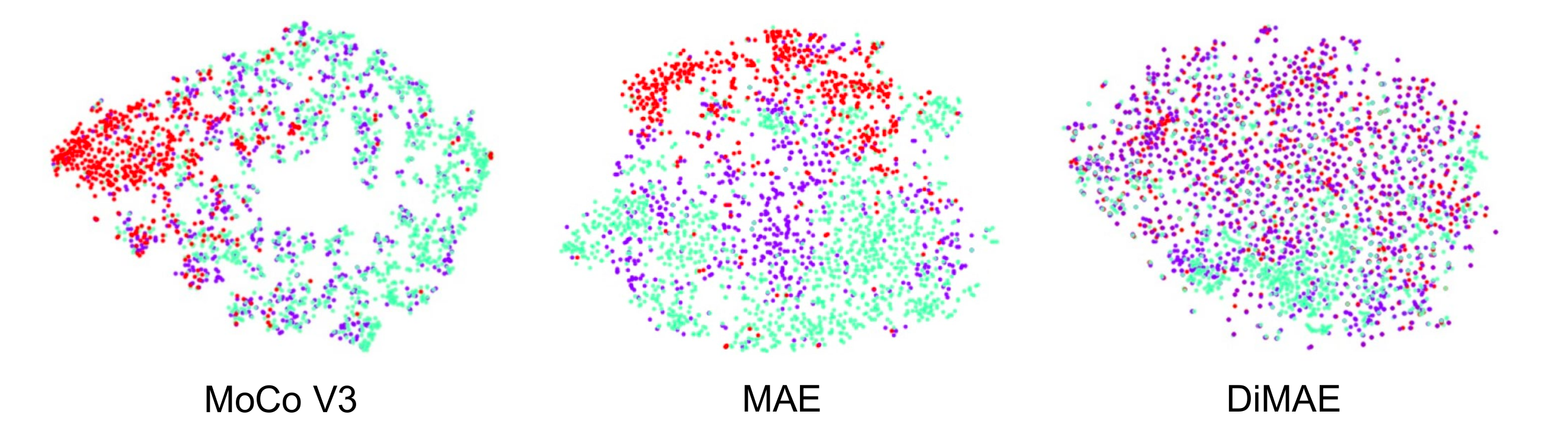}
    \caption{Visualization of the feature distribution of MoCo V3, MAE and DiMAE.}
    \label{fig:tsne}
\end{figure}

\noindent \textbf{Feature Distribution Visualization.}
Qualitatively,  Fig. \ref{fig:tsne} visualizes the feature distribution of MoCo V3, MAE and DiMAE by t-SNE, on the combination of Painting, Real, and Sketch training set in DomainNet. 
We observe that the features of DiMAE between three domains are significantly better mixed than the others. This suggests that compared with MoCo V3 and MAE, DiMAE is able to capture better domain-invariant representations.

\begin{figure}[t]
    \centering
    \footnotesize
    \includegraphics[width=0.95\linewidth]{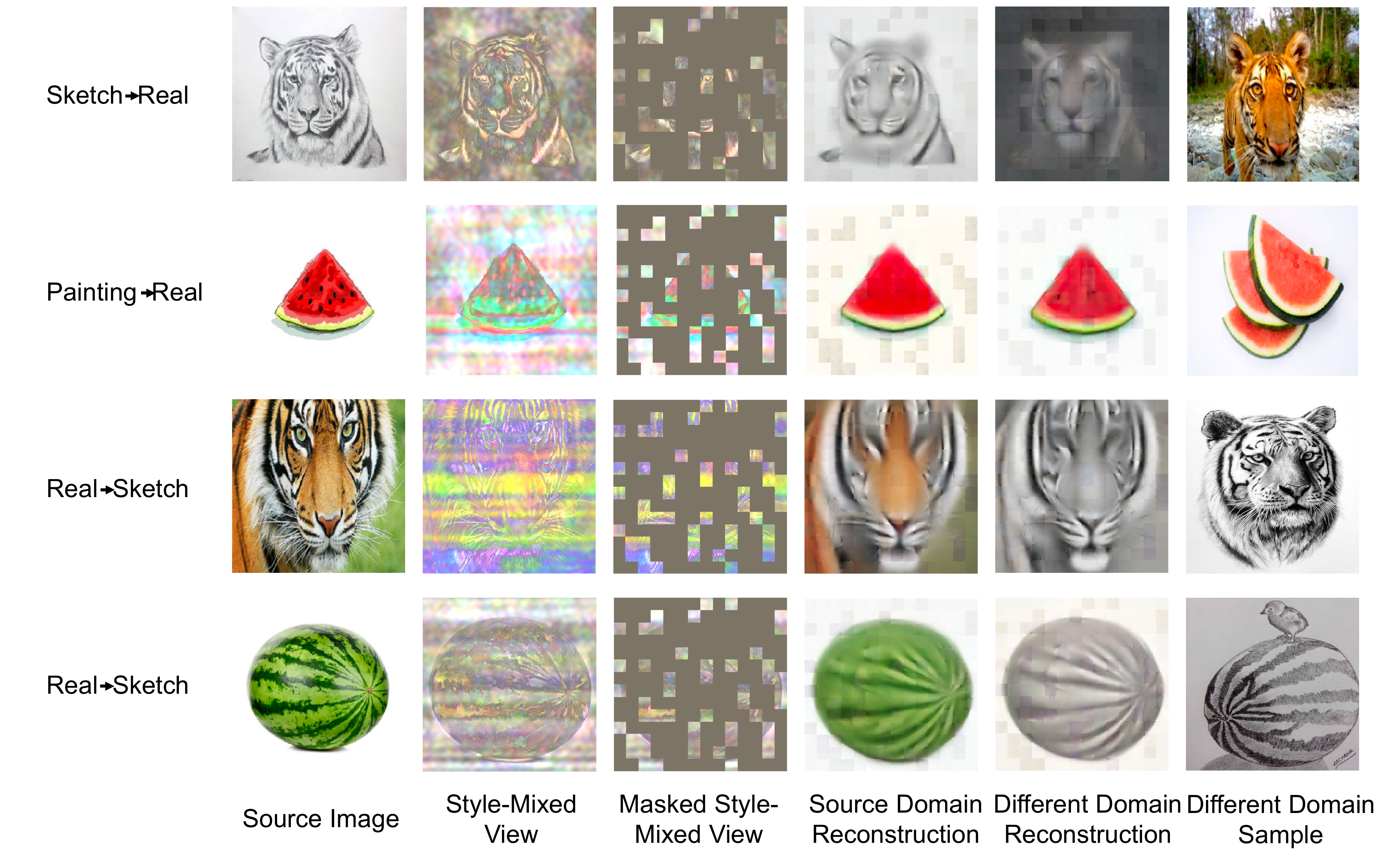}
    \caption{Reconstruction visualization of different decoders. Sketch$\rightarrow$Real denotes using Sketch as source domain and Real as the a different domain to reconstruct.}
    \vspace{-1em}
    \label{fig:vis_augs}
\end{figure}

\noindent \textbf{Reconstruction Visualization.\label{Reconstruction Visualization}} 
We visualize reconstruction results of DiMAE using ViT-base in Fig.~\ref{fig:vis_augs}. The results demonstrate that, in our DiMAE, the encoder removes the domain style and multiple decoders learn specific style information. Specifically, DiMAE eliminates the style noise on visible patches as no messy style information appears in reconstructions. Second, 
DiMAE provides complete reconstructions with specific domain styles. 
Third, we also observe that it is quite hard for DiMAE to recover colors perfectly from sketch inputs. 

\section{Conclusions}

In this paper, we propose a novel Domain invariant Masked AutoEncoder (DiMAE) to tackle the self-supervised learning from multi-domain data. Our DiMAE constructs a new cross-domain reconstruction task with a proposed content preserved style mix and multiple decoder designs to learn domain-invariant features. The content preserved style mix aims to mix style information from different domains, while preserving the image content. The multiple decoders are proposed to regularize the encoder to extract domain-invariant features. Extensive experiments validate the effectiveness of DiMAE.

\newpage
{\small
\bibliographystyle{ieee_fullname}
\bibliography{egbib}
}

\end{document}